\documentclass[letterpaper,10pt,conference]{ieeeconf}

\IEEEoverridecommandlockouts
\usepackage{amsmath}
\usepackage{amssymb}
\usepackage{cite}
\usepackage{graphicx}
\graphicspath{{figures/}}
\usepackage{url}
\usepackage{booktabs}
\usepackage{multirow}

\title{\LARGE \bf
Learning and Verifying Progress: \\ Closed-Loop Agentic Planning with Progress-Enhanced Execution \\ for Long-Horizon Vision-Language Navigation
}
\author{%
Kangjie Zhou$^{1,2,*}$, Sukai Wang$^{1,*,\dagger}$, Yuhan Xiao$^{1,*}$,
Qinghang Liu$^{2}$, Zhejia Wen$^{2}$,\\
Chunxuan Jiao$^{2}$, Wei Shan$^{1,\ddagger}$, Guanghui Ren$^{1}$,
Maoqing Yao$^{1}$, and Chang Liu$^{2,\ddagger}$%
\thanks{$^{1}$AgiBot. $^{2}$School of Advanced Manufacturing and Robotics, Peking University.}%
\thanks{$^{*}$Equal contribution. $^{\dagger}$Project leader. $^{\ddagger}$Corresponding authors.}%
}

\begin{document}

\maketitle
\thispagestyle{empty}
\pagestyle{empty}

\begin{abstract}
Long-horizon vision-language navigation (VLN) requires agents to track instruction progress and recover from local errors before they propagate across subtasks. However, executors trained on expert trajectories encounter policy-induced state shifts during deployment, and their stop signals do not reliably indicate subtask completion. We introduce Learning and Verifying Progress for VLN (LeaP-VLN), an agentic VLN system that couples progress-aware execution with independent progress verification through plan--execute--verify--recover cycles. LeaP-VLN leverages progress both explicitly, through a progress verifier that verifies subtask completion before the agent advances and triggers dual-phase recovery upon route deviation, and implicitly, by co-training the executor on progress reasoning and action prediction to enhance the executor's progress awareness and corrective control. To adapt the executor to this closed-loop interface, we propose a two-stage progress-enhanced curriculum: behavior cloning on expert demonstrations, followed by joint training on demonstrations, expert action corrections, and progress-reasoning supervision. Policy-generated trajectories expose the executor to both intra-subtask drift and inter-subtask state shifts, aligning training with the deviations encountered during closed-loop execution. Extensive experiments on R2R-CE and RxR-CE demonstrate state-of-the-art success rates among monocular methods trained without additional navigation datasets, while real-world experiments further validate the effectiveness of LeaP-VLN in long-horizon navigation.
\end{abstract}

\section{Introduction}

Vision-language navigation (VLN) requires an embodied agent to follow a natural-language route instruction in a previously unseen environment \cite{anderson2018r2r,krantz2020beyond,ku2020room}. In long-horizon settings, instructions specify sequences of landmark transitions and spatial relations. The agent must therefore predict appropriate actions while inferring from evolving egocentric observations which parts of the instruction have been completed and what remains to be executed. Moreover, local execution errors or instruction misinterpretations can compound across successive subtasks, causing increasing drift from the intended route and making subsequent stages harder to complete. Reliable long-horizon VLN requires progress tracking, accurate subtask execution, and recovery from local failures.

\begin{figure}[t]
\centering
\includegraphics[width=0.98\columnwidth]{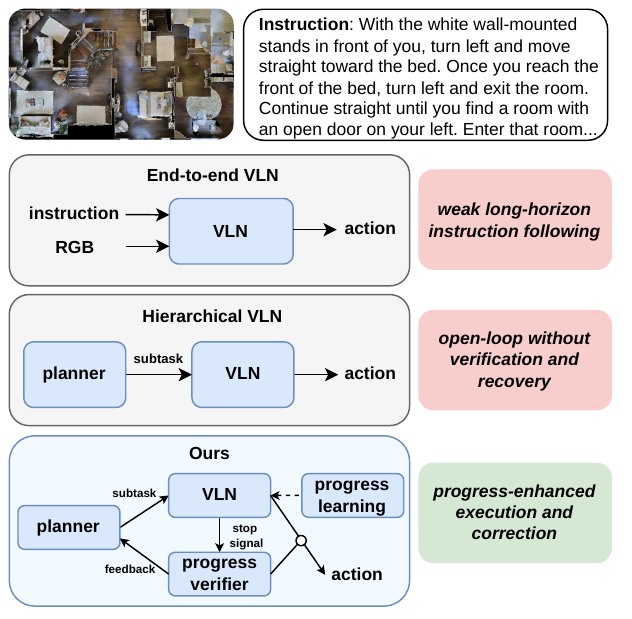}
\caption{Comparison of long-horizon VLN execution paradigms. End-to-end policies map the complete route instruction and visual observations directly to actions, with route progress implicitly inferred within the action policy. Hierarchical planner--executor pipelines without outcome verification dispatch local subtasks and advance based on executor termination. Our LeaP-VLN couples progress-aware execution with progress verification and dual-phase recovery to ground subtask transitions in observed outcomes and limit error propagation.}
\label{fig-teaser}
\end{figure}

Existing VLN approaches differ in how they organize instruction interpretation and action execution, as illustrated in Fig.~\ref{fig-teaser}. End-to-end policies built on vision-language models (VLMs) map the complete route instruction, the visual history, and the current observation directly to actions \cite{zhang2024navid,zhang2024uni,zeng2025janusvln}. Route understanding, progress tracking, and action prediction are integrated into a single policy, leaving instruction progress implicit and subtask transitions difficult to control explicitly. 
Hierarchical planner--executor systems separate high-level route reasoning from low-level control: a planner specifies target locations, waypoints, local instructions, or modular skills, and an executor carries out the corresponding navigation actions \cite{zhou2024navgpt,long2024instructnav,xin2026agentvln,zhang2026qwenrobotnavtechnicalreportscalable}. This separation makes intermediate navigation goals explicit, but reliable execution still depends on the executor's ability to handle policy-induced deviations and the planner's ability to assess subtask outcomes. Two challenges arise at this interface. First, executors trained on expert demonstrations face policy-induced state shifts during closed-loop execution, including intra-subtask drift and inter-subtask boundary shifts. Second, unverified subtask transitions can propagate local failures into inaccurate route-progress estimates. Existing recovery strategies either replan from a potentially deviated state \cite{wang2026reflectvln,xin2026agentvln} or backtrack without correcting instruction grounding \cite{shi2025smartway}. Reliable execution therefore requires handling both intra-subtask drift and inter-subtask state shifts, verifying subtask outcomes, and coupling state recovery with instruction re-grounding.

To address these gaps, we introduce LeaP-VLN, a closed-loop agentic VLN system that exploits subtask progress on both sides of the planner--executor interface (Fig.~\ref{fig-teaser}, bottom). On the planner side, progress is handled explicitly. A high-level multimodal agent decomposes the route instruction into local subtasks and orchestrates plan--execute--verify--recover cycles. An independent verifier assesses subtask completion and route alignment using the observed execution history. The agent advances only after completion is verified, and the corresponding terminal state is stored as a trusted recovery anchor. Upon route deviation, dual-phase recovery restores a trusted execution context and re-grounds the pending subtask for a refined retry. On the executor side, progress awareness is learned implicitly by co-training the executor on progress reasoning and action prediction. We propose a two-stage progress-enhanced curriculum that first initializes the executor through behavior cloning on expert demonstrations, then co-trains it using these demonstrations and expert-corrected intra- and inter-subtask policy rollouts. Learning to reason about progress content improves action prediction while promoting progress--action consistency, aligning the executor's decisions with the internal subtask progress.

Our contributions are summarized as follows.
\begin{itemize}
    \item We introduce LeaP-VLN, a closed-loop agentic navigation system for long-horizon VLN that integrates progress-aware execution, progress verification, and dual-phase recovery to maintain route alignment and limit error propagation across subtasks.
    \item We develop a two-stage progress-enhanced curriculum for training a low-level VLN executor. By co-training action prediction and progress reasoning on expert demonstrations and policy-generated trajectories, the curriculum trains the executor to handle intra-subtask drift and inter-subtask state shifts during closed-loop execution.
    \item We achieve state-of-the-art success rates on R2R-CE and RxR-CE among monocular methods trained without additional navigation datasets. Component ablations evaluate the contributions of agentic planning, progress-enhanced training, and test-time verification and recovery, while real-world experiments further validate the effectiveness of LeaP-VLN in long-horizon navigation.
\end{itemize}

\section{Related Work}

\subsection{Long-Horizon Vision-Language Navigation}
Vision-and-Language Navigation (VLN) requires embodied agents to translate language instructions and visual observations into sequential navigation actions. The task has progressed from R2R's graph-based indoor navigation to more challenging settings, with VLN-CE introducing continuous control and RxR featuring longer routes and greater linguistic diversity\cite{anderson2018r2r,krantz2020beyond,ku2020room}. 
Recent VLM-based policies directly map instructions and visual histories to actions across diverse tasks\cite{zhang2024navid,zhang2024uni,cheng2024navila,wei2025streamvln}, while generalist models broaden task coverage \cite{zhang2025embodied,gong2026abotn1} and hierarchical memory enriches visual context \cite{zeng2025janusvln}. However, these policies entangle route understanding, progress estimation, and action generation within a unified policy, leaving the currently active instruction stage implicit and allowing local execution errors to accumulate as navigation progresses, eventually leading to task failure. 

\subsection{Hierarchical Planning and Execution in VLN}
Hierarchical embodied agents decompose long-horizon tasks into high-level planning and low-level control. NavGPT adopts this paradigm for VLN by combining explicit high-level reasoning with a separate navigation controller \cite{zhou2024navgpt}. Subsequent methods extend this paradigm through local instruction dispatch, modular skill selection and waypoint prediction\cite{long2024instructnav,han2026seqwalker,wei2025ground,xin2026agentvln,zhang2026qwenrobotnavtechnicalreportscalable}. 
Despite these advances, existing methods either accept executor termination without verification, risking premature subtask transitions, or replan from a potentially deviated state, allowing errors to propagate. Consequently, local execution errors can accumulate over long-horizon navigation and ultimately lead to task failure.

\begin{figure*}[t]
    \centering
    \includegraphics[width=\textwidth]{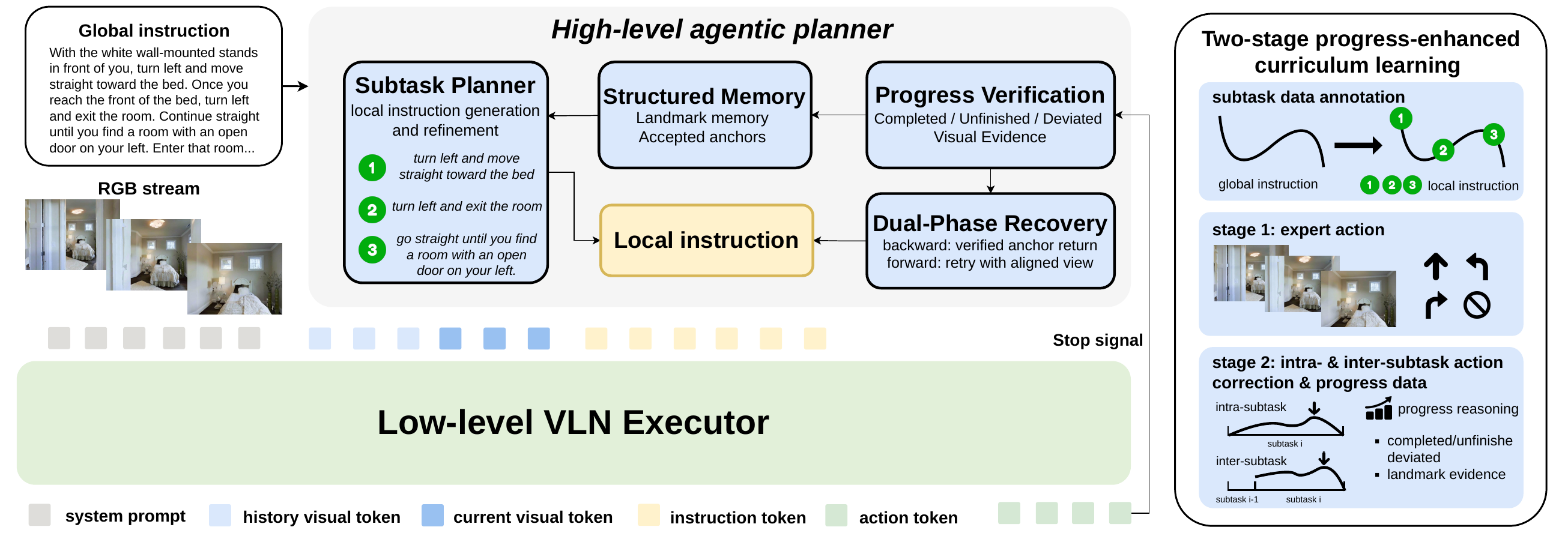}
    \caption{Overview of the proposed closed-loop agentic framework with test-time verification and recovery. Two-stage progress-enhanced curriculum learning, which clones expert actions and then co-trains intra- and inter-subtask on-policy action corrections with progress-reasoning supervision, produces the low-level VLN executor, while the high-level multimodal agent maintains landmark memory and accepted anchors and orchestrates subtask planning, progress verification, and dual-phase failure recovery.}
    \label{fig-framework}
    \end{figure*}

\subsection{Progress Monitoring and Error Recovery in VLN}
Progress estimation provides intermediate signals for tracking instruction completion, identifying navigation milestones, and determining when to terminate execution. Recent methods incorporate jointly trained semantic reasoning \cite{wang2026progressthink} or self-reflection at critical progress states \cite{guo2026awarevln}, enabling more interpretable assessments of what has been completed and what remains. However, progress assessment performed by the action policy itself allows overconfident execution errors to remain undetected. 
Beyond progress monitoring, navigation errors are addressed through policy-triggered backtracking \cite{shi2025smartway}, training-time correction \cite{yu2026correctnav}, or reflective replanning during inference \cite{wang2026reflectvln}. 
However, existing test-time mechanisms either replan from an already deviated state or backtrack without revising the instruction that caused the failure, leaving repeated errors unresolved.

\section{Problem Formulation}
\label{sec-problem}

We consider VLN in continuous environments (VLN-CE), where an agent receives a natural-language instruction $I$ at the beginning of an episode. At each step $t$, the agent observes an egocentric RGB image $o_t$ and executes a low-level action $a_t\in\{\texttt{FORWARD},\allowbreak\texttt{TURN\_LEFT},\allowbreak\texttt{TURN\_RIGHT},\allowbreak\texttt{STOP}\}$ ($0.25\,\mathrm{m}$ forward, $15^\circ$ turns).
The objective is to reach the destination specified by $I$.
 
We adopt a hierarchical formulation in which a high-level agent $\mathcal{G}$ decomposes $I$ into an ordered sequence of initial local sub-instructions,
\begin{equation}
\mathbf{S}^{(0)}=\mathcal{G}_{\mathrm{dec}}(I)=(S_1^{(0)},\ldots,S_N^{(0)}),
\label{eq-hierarchical-plan}
\end{equation}
where $\mathcal{G}_{\mathrm{dec}}$ denotes the agent's instruction-decomposition operation, $S_i^{(0)}$ specifies the initial instruction for the $i$-th subtask, and $N$ is the number of subtasks.
We use $S_i$ to denote the active local instruction dispatched to the executor, which may be adjusted from $S_i^{(0)}$ based on verified execution progress.

A low-level VLN executor $\pi_\theta$ predicts actions conditioned on the active sub-instruction $S_i$, the visual history $H_t$, and the current observation $o_t$,
\begin{equation}
a_t\sim\pi_\theta(\cdot\mid S_i,H_t,o_t).
\label{eq-exec}
\end{equation}
The executor's \texttt{STOP} signal triggers high-level progress verification, which determines whether to advance to the next subtask, continue the current subtask, or initiate recovery. The following sections detail the subtask planning, executor training, verification, and recovery mechanisms.

\section{Closed-Loop Agentic Navigation Framework}
\label{sec-framework}

As illustrated in Fig.~\ref{fig-framework}, the high-level agent $\mathcal{G}$ is a frozen multimodal model that coordinates subtask planning, progress verification, and failure recovery through a shared structured memory bank. This memory maintains trusted execution context throughout navigation. The agent invokes the executor $\pi_\theta$ as a short-horizon navigation tool within a plan--execute--verify--recover loop.

\textbf{Subtask planning with structured memory.} At the start of each episode, the agent decomposes the global instruction $I$ into an ordered sequence of local sub-instructions, each specifying a local navigation goal. This initial decomposition serves as a flexible plan rather than a fixed execution schedule. Before the initial dispatch of subtask $i$, the agent consults the updated memory $\mathcal{M}_i$ and the anchor observation $o_{t_i}$ to adjust the corresponding sub-instruction according to the verified progress,
\begin{equation}
S_i=\mathcal{G}_{\mathrm{adj}}
(S_i^{(0)},\mathcal{M}_i,o_{t_i}),\label{eq-adaptive-plan}
\end{equation}
where $\mathcal{G}_{\mathrm{adj}}$ denotes the instruction-adjustment operation performed by the high-level agent.

The memory $\mathcal{M}_i$ in Eq.~\eqref{eq-adaptive-plan} contains two components: landmark memory and accepted anchors. The landmark memory associates landmark semantic information with the corresponding visual memories acquired during validated execution. Each accepted anchor $A_i=(x_i,o_{t_i})$ records the agent's pose $x_i$ and the corresponding egocentric RGB observation $o_{t_i}$ at environment step $t_i$. The first anchor $A_1$ corresponds to the episode start state. The executor receives the active local instruction and visual inputs specified in Eq.~\eqref{eq-exec}, while the global instruction and memory are maintained by the high-level agent. This separation allows progress verification and failure handling to be performed independently for each subtask.

\textbf{Progress verification and recovery loop.} Progress verification is triggered only when the VLN executor emits a \texttt{STOP} signal. We define the decision set $\mathcal{D}=\{\mathsf{C},\mathsf{U},\mathsf{D}\}$, where $\mathsf{C}$, $\mathsf{U}$, and $\mathsf{D}$ denote completed, unfinished but on-route, and deviated, respectively, and the verifier assigns an outcome $d_i\in\mathcal{D}$. For $d_i=\mathsf{C}$, the landmark memory is updated with the verified landmark semantics and visual evidence, the terminal state is committed as the next anchor $A_{i+1}$, and the planner advances to the next subtask if $i<N$; otherwise, the episode terminates. For $d_i=\mathsf{U}$, the landmark memory is updated with the confirmed progress, and the planner issues a continuation instruction for the unfinished portion of the current subtask. The executor follows this instruction from the current state. For $d_i=\mathsf{D}$, the memory remains unchanged, and dual-phase recovery returns the agent to $A_i$ and uses active perception to select a suitable retry view. The executor then retries the pending subtask with its visual history reset. Sec.~\ref{sec-executor} describes how the executor is trained for this interface, while Sec.~\ref{sec-recovery} details verification and recovery.

\section{Low-Level VLN Executor Training}
\label{sec-executor}

To enable robust local instruction following under policy-induced deviations, we develop the training pipeline (Fig.~\ref{fig-training}) comprising boundary-aligned supervision construction from route-level annotations (Fig.~\ref{fig-training}(a)) and a two-stage progress-enhanced curriculum for co-training the executor on action prediction and progress reasoning (Fig.~\ref{fig-training}(b)).

\subsection{Local-Instruction Data Generation}

\begin{figure}[t]
\centering
\includegraphics[width=\columnwidth]{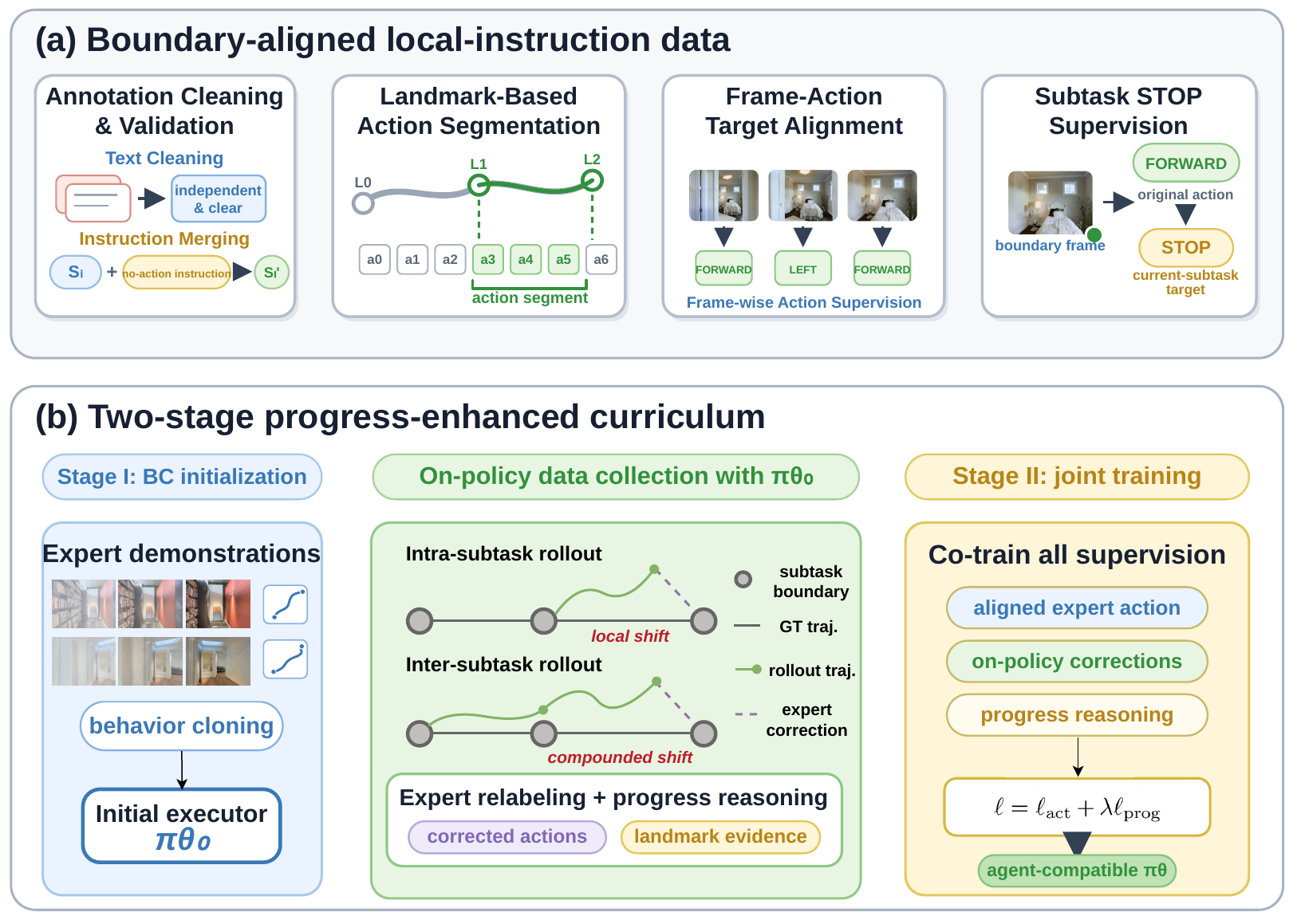}
\caption{Low-level VLN executor training. (a) Boundary-aligned local-instruction data generation cleans and merges segment annotations, aligns frame--action targets, and adds explicit \texttt{STOP} supervision at subtask boundaries. (b) The two-stage progress-enhanced curriculum first initializes the executor by behavior cloning, then collects intra- and inter-subtask on-policy rollouts for expert action correction and progress reasoning, and finally jointly trains on all supervision.}
\label{fig-training}
\end{figure}

Under the agentic interface, we reorganize RxR route-level demonstrations into subtask-level samples using the aligned sub-instructions and sub-paths provided by Landmark-RxR \cite{he2021landmarkrxr}. As shown in Fig.~\ref{fig-training}(a), we first clean and validate the annotations by merging trailing no-action instructions into the preceding valid instruction, and retaining only routes whose sub-paths form a complete and continuous trajectory. 
We then use the annotated landmarks to partition each expert action sequence into subtask segments. During replay in Habitat, we align graph-level segment boundaries with continuous observations and actions, convert actions into next-action targets, and construct horizon-$K$ chunks that do not cross subtask boundaries. At each boundary, the terminal observation is assigned an explicit \texttt{STOP} target instead of the original route-continuation action and is reused as the initial context for the next subtask.

\subsection{Two-Stage Progress-Enhanced Curriculum Learning}

Boundary-aligned demonstrations train the executor to follow local instructions from expert states, but provide limited coverage of policy-induced states and no explicit supervision for assessing subtask progress. As shown in Fig.~\ref{fig-training}(b), our curriculum first trains an initial policy through behavior cloning and then rolls it out to collect states exhibiting intra-subtask drift and inter-subtask boundary shift. We next generate progress-reasoning supervision for these states and finally co-train the executor on action prediction and progress reasoning using expert demonstrations and rollout-derived supervision.

\subsubsection{Stage I: Policy Initialization with Behavior Cloning}

In the first stage, we obtain an initial executor $\pi_{\theta_0}$ through behavior cloning on the expert demonstrations comprising full-route R2R samples and boundary-aligned local-instruction samples from RxR. 
Following Eq.~\eqref{eq-exec}, the executor predicts a horizon-$K$ action chunk $y_t=(a_t,\ldots,a_{t+K-1})$ from the context $c_t=(S_i,H_t,o_t)$, where $H_t$ contains up to $K_h$ recent frames from the current attempt. The action loss is
\[
\ell_{\mathrm{act}}=
-\sum_{j=0}^{K-1}\log p_\theta
(a_{t+j}\mid y_{t,<j},c_t).
\]

\subsubsection{Intra- and Inter-Subtask Rollouts for Action Correction and Progress Reasoning}

We then roll out $\pi_{\theta_0}$ under local instructions and relabel its visited states with expert corrective action chunks \cite{ross2011dagger}. Intra-subtask rollouts start from a ground-truth subtask boundary and expose deviations accumulated before the corresponding ground-truth endpoint. Inter-subtask rollouts instead start the next instruction from the previous rollout's actual endpoint, thereby exposing the shifted initial states inherited across subtask boundaries. For states visited in both settings, expert relabeling supplies actions that return execution toward the reference trajectory.

We further use these rollouts to construct progress-reasoning data, strengthening the executor's ability to understand navigation progress and judge the current subtask state. Specifically, for each rollout context $c_t=(S_i,H_t,o_t)$, whose visual input consists of the current observation $o_t$ and historical frames $H_t$, we construct a structured target
\[
\begin{aligned}
g_t&=(d_t,r_t),&
d_t&\in\mathcal{D},\\
r_t&=(r_t^{\mathrm{done}},r_t^{\mathrm{remain}},
r_t^{\mathrm{dev}}).
\end{aligned}
\]
The accompanying natural-language reasoning $r_t$ grounds the decision in the visual sequence: $r_t^{\mathrm{done}}$ describes which part of the local instruction has already been completed, $r_t^{\mathrm{remain}}$ identifies the unmet progress and completion condition, and $r_t^{\mathrm{dev}}$ analyzes whether the trajectory has deviated and cites the corresponding visual evidence. Thus, each policy-induced observation sequence is paired with both a corrected action target and an explicit progress decision with structured reasoning; segment annotations are used only to generate this training supervision.

\subsubsection{Stage II: Joint Progress-Enhanced Training}

In the second stage, we co-train the executor on the aligned subtask demonstrations, expert-corrected samples from both rollout settings, and the structured progress-reasoning targets $g_t$. The joint objective is
\[
\ell = \ell_{\mathrm{act}} + \lambda \ell_{\mathrm{prog}},
\qquad
\ell_{\mathrm{prog}}=
-\sum_{k}\log p_\theta(g_{t,k}\mid g_{t,<k},c_t),
\]
where $\ell_{\mathrm{act}}$ is computed over both demonstration and correction samples, and $\lambda$ balances action learning and progress learning. The resulting low-level executor combines action prediction with progress-aware reasoning, enabling accurate subtask execution and reliable transitions across subtask boundaries.

\section{Progress Verification and Failure Recovery}
\label{sec-recovery}

Although progress-reasoning supervision strengthens the executor's ability to assess subtask progress, its stop signal does not guarantee that the current subtask has been successfully completed. We therefore trigger progress verification before any memory update. The verification result determines whether the agent advances to the next subtask, continues the unfinished current subtask, or enters dual-phase recovery.

\subsection{Progress Verification and Memory Update}

A progress verification is requested when the executor outputs a stop signal. At the proposed termination time $t$, the verifier follows the same input structure as the executor, receiving the current local instruction $S_i$, the observation history $H_t$, and the current observation $o_t$. Here, $H_t$ contains historical frames sampled from the current subtask attempt, while $o_t$ is the final frame at which the stop signal is emitted. The verifier judges both route alignment and subtask completion, producing
\[
J_i=\mathcal{V}(S_i,H_t,o_t),
\]
where $J_i$ contains a three-way decision $d_i\in\mathcal{D}$ together with its supporting evidence.

The decision controls both execution and memory updates. If $d_i=\mathsf{C}$, the verified landmark semantics and supporting frames are added to the landmark memory, while the terminal state is independently appended as the next anchor $A_{i+1}$. The resulting memory $\mathcal{M}_{i+1}$ supports planning for $S_{i+1}$; completion of the final segment terminates the episode. If $d_i=\mathsf{U}$, the landmark memory is updated with the confirmed progress, and the agent then continues the unfinished portion of $S_i$ from the current state. If $d_i=\mathsf{D}$, the memory remains unchanged and dual-phase recovery is invoked from the latest accepted anchor.

\subsection{Dual-Phase Test-Time Recovery}

For $d_i=\mathsf{D}$, the agent has already deviated from the intended route, making forward-only recovery from the corrupted execution state unreliable. We therefore introduce dual-phase recovery, consisting of backward and forward phases, before retrying the failed subtask. In the backward phase, the agent returns to the latest accepted anchor $A_i$, thereby restoring the validated execution state from which $S_i$ was originally dispatched.
In the forward phase, the agent performs active perception by acquiring a panorama $P_i$ at the restored anchor. Conditioned on $S_i$, the verification evidence $J_i$, and $P_i$, it selects the most suitable view $v_i^*$ for the next execution attempt. The agent then aligns itself with $v_i^*$ and retries the pending subtask with a reset executor history, grounding the new attempt in the most informative local view. By jointly restoring a validated state and re-grounding execution through active perception, this procedure reduces ambiguity at retry initialization and prevents local deviations from propagating to subsequent subtasks.

\section{Experiments}

\begin{table*}[t]
\centering
\caption{Comparison on the R2R-CE and RxR-CE val-unseen splits.}
\label{tab-results}
\scriptsize
\setlength{\tabcolsep}{2pt}
\renewcommand{\arraystretch}{1.02}
\begin{tabular*}{\textwidth}{@{\extracolsep{\fill}}l cccc cccc ccc@{}}
\toprule
\multirow{2}{*}{Method} & \multicolumn{4}{c}{Observation} & \multicolumn{4}{c}{R2R-CE Val-Unseen} & \multicolumn{3}{c}{RxR-CE Val-Unseen} \\
\cmidrule(lr){2-5} \cmidrule(lr){6-9} \cmidrule(lr){10-12}
 & S.RGB & Pano. & Depth & Odom. & NE$\downarrow$ & OSR$\uparrow$ & SR$\uparrow$ & SPL$\uparrow$ & NE$\downarrow$ & SR$\uparrow$ & SPL$\uparrow$ \\
\midrule
CMA~\cite{krantz2020beyond} & & \checkmark & \checkmark & \checkmark & 6.20 & 52.0 & 41.0 & 36.0 & 8.76 & 26.5 & 22.1 \\
InstructNav~\cite{long2024instructnav} & & \checkmark & \checkmark & \checkmark & 6.89 & -- & 31.0 & 24.0 & -- & -- & -- \\
ETPNav~\cite{an2024etpnav} & & \checkmark & \checkmark & \checkmark & 4.71 & 65.0 & 57.0 & 49.0 & 5.64 & 54.7 & 44.8 \\
HNR~\cite{wang2024lookahead} & & \checkmark & \checkmark & \checkmark & 4.42 & 67.0 & 61.0 & 51.0 & 5.50 & 56.3 & 46.7 \\
NavFoM~\cite{zhang2025embodied} & & \checkmark & & & 4.61 & 72.1 & 61.7 & 55.3 & 4.74 & 64.4 & 56.2 \\
ABot-N1~\cite{gong2026abotn1} & & \checkmark & & & 3.91 & 71.7 & 68.3 & 66.6 & 3.43 & 70.9 & 61.4 \\
Qwen-RobotNav-4B~\cite{zhang2026qwenrobotnavtechnicalreportscalable} & & \checkmark & & & 3.80 & 77.2 & 69.5 & 63.6 & 3.80 & 75.2 & 65.0 \\
\midrule
LAW~\cite{raychaudhuri2021language} & \checkmark & & \checkmark & \checkmark & 6.83 & 44.0 & 35.0 & 31.0 & 10.90 & 8.0 & 8.0 \\
CM$^2$~\cite{georgakis2022crossmodal} & \checkmark & & \checkmark & \checkmark & 7.02 & 41.0 & 34.0 & 27.0 & -- & -- & -- \\
WS-MGMap~\cite{chen2022wsmgmap} & \checkmark & & \checkmark & \checkmark & 6.28 & 47.0 & 38.0 & 34.0 & -- & -- & -- \\
Seq2Seq~\cite{krantz2020beyond} & \checkmark & & \checkmark & & 7.77 & 37.0 & 25.0 & 22.0 & 12.10 & 13.9 & 11.9 \\
AgentVLN~\cite{xin2026agentvln} & \checkmark & & \checkmark & & 3.88 & 73.5 & 67.2 & 64.7 & 3.92 & 69.5 & 61.3 \\
\midrule
NaVid~\cite{zhang2024navid} & \checkmark & & & & 5.47 & 49.0 & 37.0 & 35.0 & -- & -- & -- \\
Uni-NaVid$^\dagger$~\cite{zhang2024uni} & \checkmark & & & & 5.58 & 53.5 & 47.0 & 42.7 & 6.24 & 48.7 & 40.9 \\
NaVILA$^\dagger$~\cite{cheng2024navila} & \checkmark & & & & 5.22 & 62.5 & 54.0 & 49.0 & 6.77 & 49.3 & 44.0 \\
StreamVLN$^\dagger$~\cite{wei2025streamvln} & \checkmark & & & & 4.98 & 64.2 & 56.9 & 51.9 & 6.22 & 52.9 & 46.0 \\
DualVLN$^\dagger$~\cite{wei2025ground} & \checkmark & & & & \underline{4.05} & 70.7 & 64.3 & 58.5 & 4.58 & 61.4 & 51.8 \\
CorrectNav~\cite{yu2026correctnav} & \checkmark & & & & 4.24 & 67.5 & 65.1 & \textbf{62.3} & \underline{4.09} & 69.3 & \textbf{63.3} \\
Qwen-RobotNav-4B$^\dagger$~\cite{zhang2026qwenrobotnavtechnicalreportscalable} & \checkmark & & & & 4.22 & \underline{73.6} & \underline{66.9} & \underline{60.5} & 4.15 & \textbf{71.3} & \underline{61.5} \\
\textbf{LeaP-VLN (Ours)} & \checkmark & & & & \textbf{3.71} & \textbf{73.9} & \textbf{67.5} & 59.1 & \textbf{3.89} & \underline{69.6} & 52.8 \\
\bottomrule
\end{tabular*}
\vspace{2pt}

\begin{minipage}{\textwidth}
\scriptsize
S.RGB, Pano., Depth, Odom.: monocular RGB, panoramic observations, depth, and odometry inputs. $\dagger$: trained with additional navigation datasets. Bold/underline: best/second best within the RGB-only observation group.
\end{minipage}
\end{table*}

\subsection{Experimental Setup}

To validate the effectiveness of our method, we evaluate on the R2R-CE \cite{krantz2020beyond} and RxR-CE \cite{ku2020room} benchmarks in VLN-CE. We organize VLN-CE baselines by their observation inputs: panoramic methods, monocular RGB augmented with additional modalities such as depth, and monocular RGB alone, and primarily compare our framework with methods under the same RGB-only observation setting. We report standard VLN-CE metrics, including Success Rate (SR), Success weighted by Path Length (SPL), Oracle Success Rate (OSR), and Navigation Error (NE).

\subsection{Implementation Details}

We employ Qwen-3.7-Plus as the high-level agent and initialize the low-level executor from Qwen3.5-VL-4B-Instruct. In the first training stage, the executor is trained exclusively on 1.4M expert action samples, comprising R2R samples with complete route instructions and subtask-level samples obtained by decomposing RxR trajectories. The second stage uses the full set of 2.2M action samples, combining these expert-derived samples with 0.8M samples labeled with corrective actions, together with approximately 200K progress-reasoning samples. At each decision step, the executor conditions on the current observation and up to eight historical frames, using all available preceding frames when fewer than eight exist, to predict a chunk of four action tokens. Following a receding-horizon execution scheme, only the first two predicted actions are executed before the action chunk is regenerated. The progress verifier is likewise implemented using Qwen-3.7-Plus.

\subsection{Result Analysis}

\begin{figure}[t]
\centering
\includegraphics[width=0.9\columnwidth]{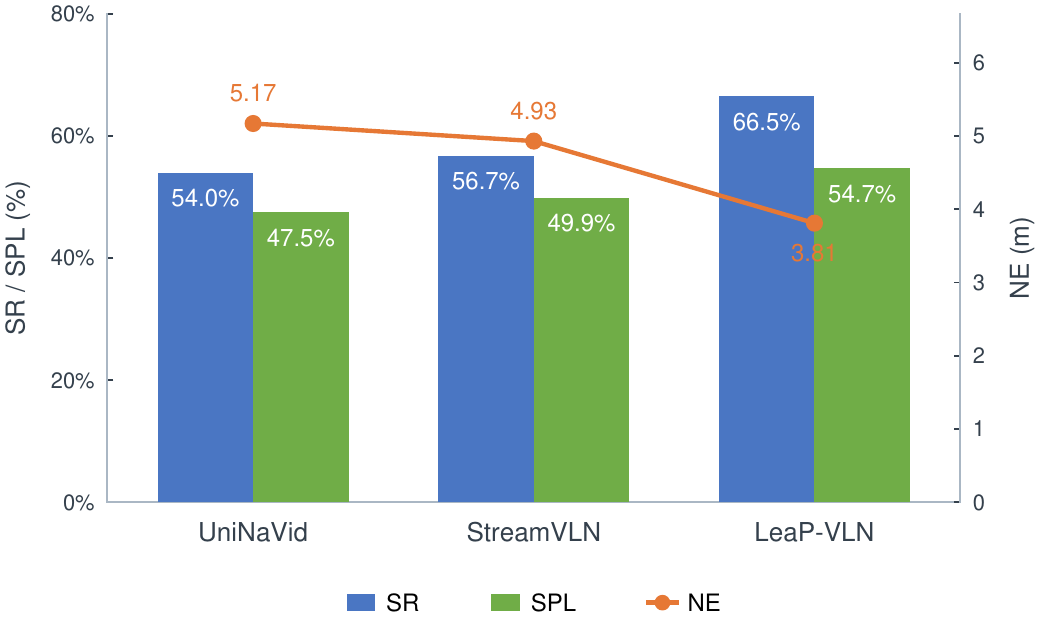}
\caption{Comparison with Uni-NaVid and StreamVLN on hard episodes (ground-truth action sequences exceed 50 steps) in VLN-CE.}
\label{fig-long-horizon}
\end{figure}

Tab.~\ref{tab-results} compares our framework with prior VLN-CE methods on the R2R-CE and RxR-CE val-unseen splits. On R2R-CE, our framework achieves the best NE, OSR, and SR among monocular RGB methods trained without additional navigation datasets. Compared with the strongest prior results under the same setting, it increases SR from 65.1\% to 67.5\% and OSR from 67.5\% to 73.9\%, while reducing NE from 4.24\,m to 3.71\,m. On the longer RxR-CE benchmark, our method achieves the best NE and SR among monocular methods trained without additional navigation datasets. When methods trained with additional navigation data are included, our framework ranks second in SR, behind only Qwen-RobotNav~\cite{zhang2026qwenrobotnavtechnicalreportscalable}. Collectively, these results demonstrate the effectiveness of our framework in improving goal-reaching reliability and reducing navigation error without relying on additional navigation data. 
Despite these gains, our framework records lower SPL than the strongest baselines on both benchmarks because anchor return and refined retries can lengthen the executed trajectory. This loss in path efficiency is an acceptable trade-off for more reliable completion in long-horizon navigation. 
As shown in Fig.~\ref{fig-long-horizon}, on hard episodes in VLN-CE whose ground-truth action sequences exceed 50 steps, LeaP-VLN substantially outperforms two representative open-source models, Uni-NaVid~\cite{zhang2024uni} and StreamVLN~\cite{wei2025streamvln}, confirming a clear advantage on long-horizon VLN tasks.

\begin{figure*}[t]
    \centering
    \includegraphics[page=1,width=0.98\textwidth]{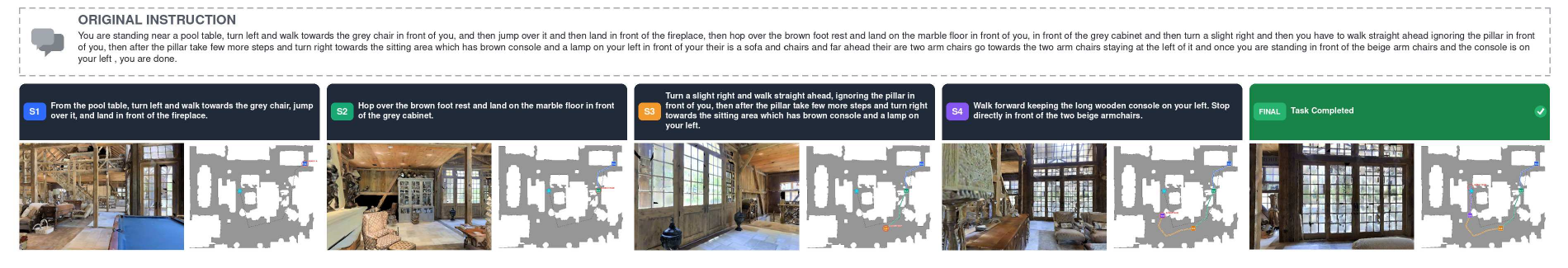}\\
    \includegraphics[page=2,width=0.98\textwidth]{VLN_qualitative_episodes_690_6505.pdf}
    \caption{Qualitative visualization of long-horizon instruction following. The complete navigation instruction is shown at the top of each example. For every subtask, the egocentric observation is paired with the corresponding top-down map, where the colored trajectory indicates the agent's stage-wise navigation progress. In the top example, the agent sequentially executes all subtasks and successfully reaches the target location. In the bottom example, the verifier detects a mid-route deviation and triggers a subtask retry, enabling successful completion.}
    \label{fig-case-study}
    \end{figure*}

Figure~\ref{fig-case-study} presents two representative long-horizon episodes. In the top example, the agent sequentially executes the decomposed subtasks and registers verified terminal states as anchors until reaching the target. In the bottom example, the verifier detects a deviation and triggers a refined retry from the latest anchor, restoring route alignment. Together, the examples show that subtask decomposition structures long-horizon execution, while progress-aware verification closes the planning-execution loop and prevents local errors from propagating.

\subsection{Ablation Studies}

\begin{table}[t]
    \centering
    \caption{Ablation of the proposed framework components on the R2R-CE and RxR-CE val-unseen splits.}
    \label{tab-ablation}
    \scriptsize
    \setlength{\tabcolsep}{2pt}
    \resizebox{0.8\columnwidth}{!}{%
    \begin{tabular}{ccccccccccc}
    \toprule
    \multicolumn{4}{c}{Components} & \multicolumn{4}{c}{R2R-CE Val-Unseen} & \multicolumn{3}{c}{RxR-CE Val-Unseen} \\
    \cmidrule(lr){1-4} \cmidrule(lr){5-8} \cmidrule(lr){9-11}
    A & B & C & D & NE$\downarrow$ & OSR$\uparrow$ & SR$\uparrow$ & SPL$\uparrow$ & NE$\downarrow$ & SR$\uparrow$ & SPL$\uparrow$ \\
    \midrule
    \checkmark & & & & 4.96 & 64.8 & 55.5 & 49.7 & 5.78 & 58.6 & 42.3 \\
    \checkmark & \checkmark & & & 4.46 & 69.3 & 61.7 & 52.4 & 4.33 & 65.2 & 51.5 \\
     & \checkmark & \checkmark & & \underline{3.78} & \underline{73.0} & \underline{67.0} & \underline{59.0} & 4.86 & 62.5 & 50.1 \\
    \checkmark & \checkmark & \checkmark & & 3.91 & 71.7 & 66.7 & 57.6 & \underline{4.10} & \underline{67.9} & \textbf{54.4} \\
    \checkmark & \checkmark & \checkmark & \checkmark & \textbf{3.71} & \textbf{73.9} & \textbf{67.5} & \textbf{59.1} & \textbf{3.89} & \textbf{69.6} & \underline{52.8} \\
    \bottomrule
    \end{tabular}%
    }
    \vspace{2pt}
    
    \begin{minipage}{\columnwidth}
    \scriptsize
    \raggedright
    A: agentic planning; B: second-stage action correction; C: progress co-training; D: progress verification and recovery.
    \end{minipage}
    \end{table}

\begin{figure}[t]
\centering
\includegraphics[width=\columnwidth]{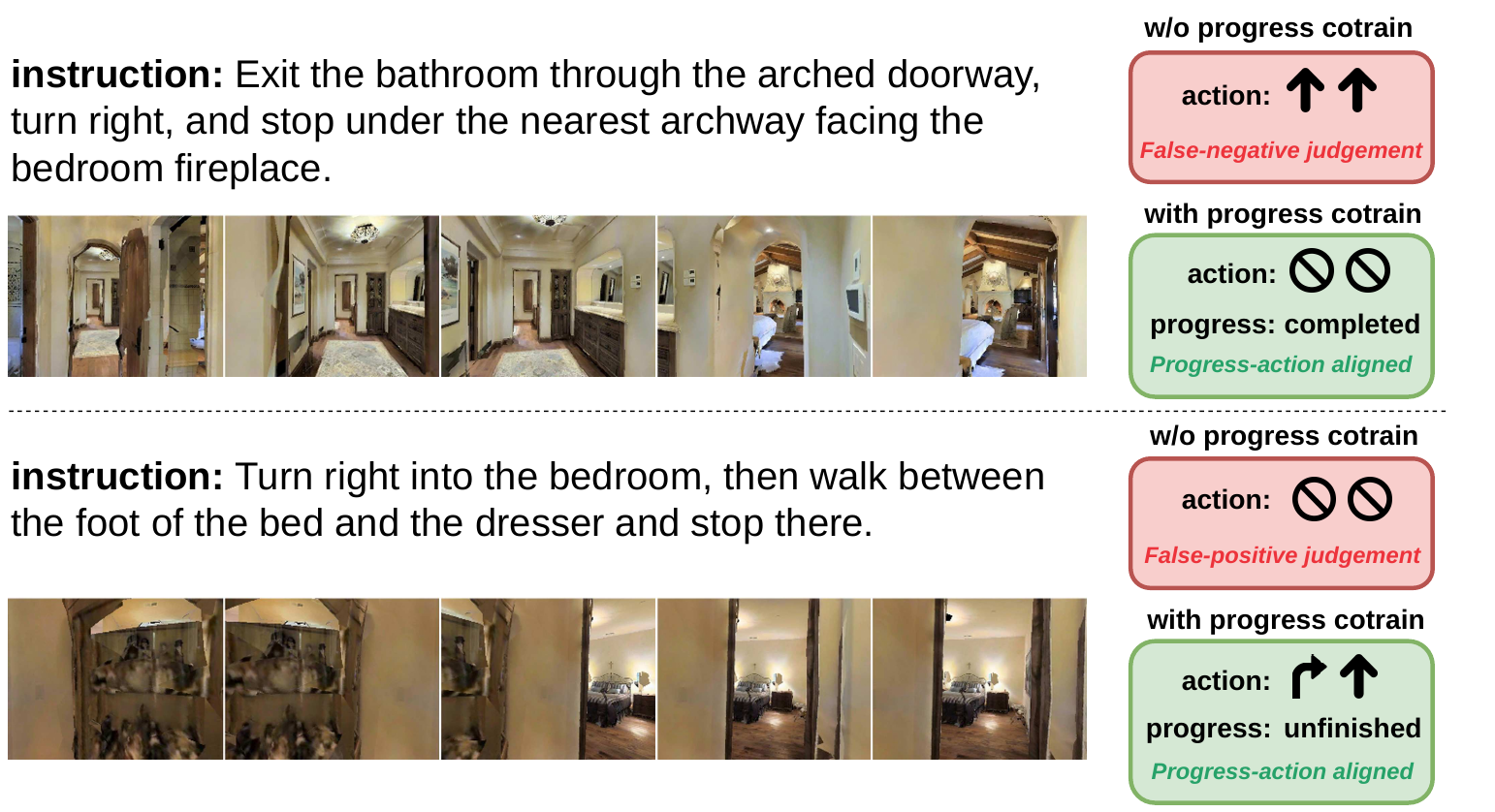}
\caption{Qualitative effects of progress co-training on action prediction. Without co-training, the executor either continues after task completion (top) or stops prematurely (bottom). Co-training improves action-progress consistency, producing \texttt{STOP} for completed subtasks and continued execution for unfinished ones.}
\label{fig-progress-cotrain}
\end{figure}

We evaluate different combinations of the proposed components on the R2R-CE and RxR-CE val-unseen splits to isolate the effectiveness of each component.
As shown in Tab.~\ref{tab-ablation}, action correction and progress co-training consistently improve SR and reduce NE on both benchmarks, while progress verification and recovery further enable the complete model to achieve the best SR and NE. The lower RxR-CE SPL after adding recovery stems from the additional travel introduced by anchor return and corrective retries.
Agentic planning yields a clear advantage on the longer RxR-CE benchmark, whereas the B+C configuration performs slightly better on R2R-CE. This is because R2R action training uses complete, undecomposed route instructions and its routes are generally shorter and simpler, making direct execution more effective. In this setting, the additional coupling among planning, execution, verification, and recovery creates more intermediate failure points whose errors can propagate through the pipeline and offset the benefit of explicit decomposition.

Fig.~\ref{fig-progress-cotrain} further illustrates the role of progress co-training. Without the co-training, the executor may continue navigating after reaching the target, indicating a false-negative estimate of task completion, or terminate before fully satisfying the instruction, indicating a false-positive estimate. By jointly learning progress estimation and action prediction, progress co-training promotes consistency between the inferred progress and generated actions, thereby mitigating both target overshooting and premature termination.

\subsection{Real-World Experiments}

\begin{figure*}[t]
    \centering
    \includegraphics[width=0.95\textwidth]{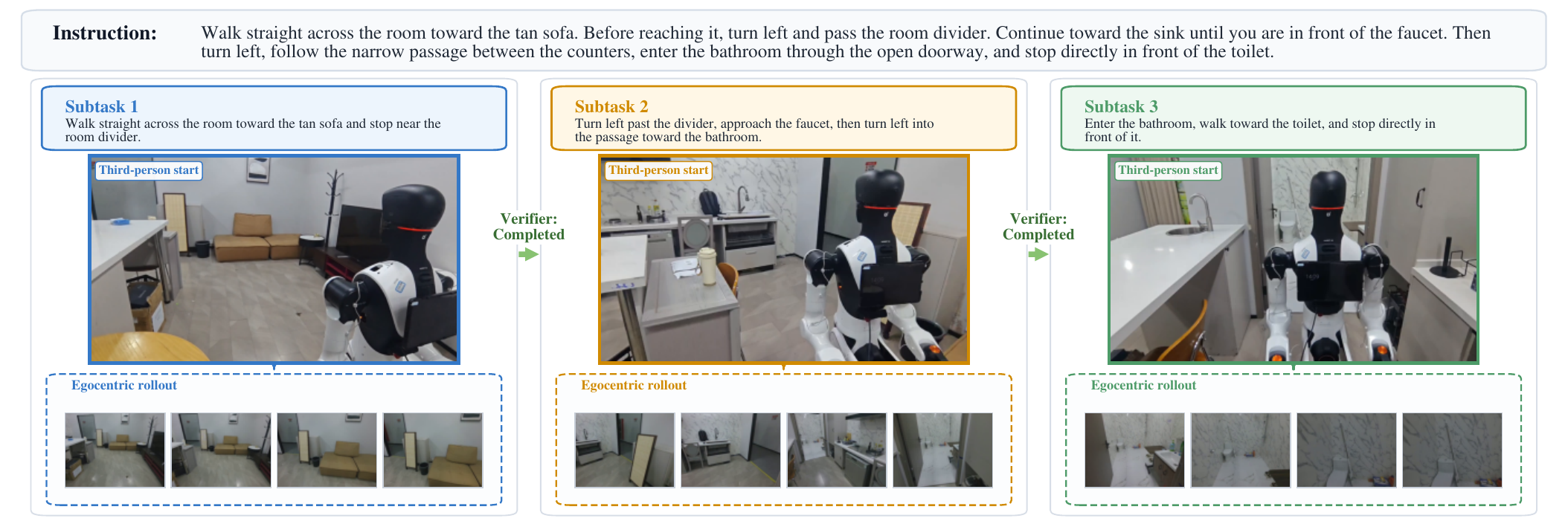}
    \caption{Real-world long-horizon navigation on the AgiBot Genie 02 robot. LeaP-VLN decomposes the route instruction into three local subtasks: approaching the divider, passing the sink, and stopping before the toilet. For each subtask, the top image shows the robot's third-person start state and the bottom row shows the corresponding egocentric image.}
    \label{fig-real-world}
\end{figure*}

To evaluate sim-to-real transfer, we deploy LeaP-VLN on an AgiBot Genie 02 robot with an RGB camera in a real-world domestic environment. We directly employ the agent, executor, and verifier used in simulation without any additional fine-tuning. Predicted actions are translated into low-level motion commands, while anchor return is executed through SLAM-based point navigation.
Fig.~\ref{fig-real-world} presents a successful episode in which the agent decomposes the route into three grounded subtasks: approaching the room divider, passing the sink, and stopping before the toilet. Third-person views show the initial state of each subtask, while the corresponding egocentric rollouts illustrate its stage-wise execution. This qualitative result demonstrates the ability of LeaP-VLN to maintain progress-consistent execution across scene transitions during real-world long-horizon navigation.

\section{Conclusion}

We presented LeaP-VLN, a closed-loop agentic system that exploits subtask progress across the planner--executor interface for long-horizon vision-language navigation. On the planner side, a multimodal agent decomposes route instructions into local subtasks, while a progress verifier checks subtask completion and route alignment before progression. Upon deviation, dual-phase recovery restores a trusted anchor and re-grounds the pending subtask. On the executor side, a two-stage progress-enhanced curriculum combines behavior cloning with joint action and progress-reasoning supervision, improving robustness to intra-subtask drift and inter-subtask state shifts. LeaP-VLN achieves state-of-the-art success rates on R2R-CE and RxR-CE among monocular methods trained without additional navigation data. Real-world experiments further validate the effectiveness of LeaP-VLN in complex navigation tasks.

\bibliographystyle{IEEEtran}
\bibliography{icra2027}

\end{document}